\documentclass[conference]{IEEEtran}
\IEEEoverridecommandlockouts

\usepackage{booktabs}
\usepackage{algorithm,algorithmic}
\usepackage{graphicx,color}

\usepackage{multirow}
\usepackage{float}
\usepackage{tabularx}
\usepackage{cuted}
\usepackage{xspace}
\usepackage[USenglish,american]{babel}
\usepackage{placeins}
\usepackage{diagbox}
\usepackage{subcaption}
\usepackage{nameref}
\usepackage{mathtools}
\usepackage{stmaryrd}
\usepackage{cleveref}
\usepackage{url}

\usepackage{cite}
\usepackage{amsmath,amssymb,amsfonts}
\usepackage{algorithmic}
\usepackage{textcomp}
\usepackage{xcolor}
\def\BibTeX{{\rm B\kern-.05em{\sc i\kern-.025em b}\kern-.08em
    T\kern-.1667em\lower.7ex\hbox{E}\kern-.125emX}}
\begin{document}

\newcommand{\ad}[2]{{\color{gray}#1}{\color{red}#2}}
\newcommand{\abm}[2]{{\color{gray}#1}{\color{orange}#2}}
\newcommand{\at}[2]{{\color{gray}#1}{\color{purple}#2}}
\newcommand{\gf}[2]{{\color{gray}#1}{\color{cyan}#2}}
\newcommand{\rv}[2]{{\color{gray}#1}{\color{green}#2}}
\newcommand{\nada}[1]{}
\newcolumntype{H}{>{\setbox0=\hbox\bgroup}c<{\egroup}@{}}

\title{Generalization Limits in Vehicle Re-Identification \\

\thanks{} 
}

\author{
\IEEEauthorblockN{
A. Y. Ben Mabrouk\textsuperscript{1} \quad
A. Tadros\textsuperscript{1} \quad
R. Grompone von Gioi\textsuperscript{1} \quad
G. Facciolo\textsuperscript{1,3} \quad
A. Davy\textsuperscript{2} \quad
R. Verschae\textsuperscript{4}
}
\IEEEauthorblockA{
\textsuperscript{1}\textit{Université Paris-Saclay, ENS Paris-Saclay, CNRS, Centre Borelli}
\quad
\textsuperscript{2}\textit{HGH Infrared Systems} \\
\textsuperscript{3}\textit{Institut Universitaire de France} 
\quad
\textsuperscript{4}\textit{Universidad Técnica Federico Santa María, Chile}
}
\thanks{Corresponding author: anis.ben\_mabrouk@ens-paris-saclay.fr}
}

\maketitle

\begin{abstract}

Vehicle re-identification focuses on retrieving images of the same vehicle from a gallery given a query image. Upon closer inspection of commonly used datasets, we observe that vehicles with few visual differences—e.g., the same make, model, and color—appear in both the training and test sets. As a result, methods that effectively memorize the training data tend to perform well on these test sets but struggle to generalize to other datasets. In this paper, we address this issue by proposing a novel evaluation approach that more effectively measures generalization capability to unseen vehicle types. To further study generalization performance, we also propose splitting the evaluation based on view, allowing us to differentiate the effect of viewpoint robustness from that of same-view re-identification. Our findings reveal that most state-of-the-art methods struggle with unseen vehicle types, and that their robustness to viewpoint changes and attention to detail are limited to vehicle types seen during training.
\end{abstract}


\section{Introduction}
In computer vision, both recognition and re-identification consist of determining if two images pertain to the same instance of an object while ideally being robust to changes like pose~\cite{5204071}, view-point changes~\cite{Sun_2019_CVPR}, modality~\cite{li2020infrared}, occlusions~\cite{steger2002occlusion}, and degradations~\cite{8283763,koo2022survey,wang2019survey}. Objects can be anything from faces~\cite{wang2021deep} people~\cite{ye2021deep},   animals~\cite{ye2024transformerobjectreidentificationsurvey,Cermak_2024_WACV} to vehicles~\cite{khan2019survey}. 
In vehicle re-identification, the objects of interest are rigid unlike faces or animals that can vary with expression, pose or age. However, vehicles of the same make appear nearly identical across identities, making them harder to distinguish especially since changes in viewpoint can still produce substantial visual differences.
Hence, the vehicle re-identification setting is well-suited for studying viewpoint robustness and fine-grained detail recognition.

\begin{figure}[t]
    \centering
    \resizebox{\linewidth}{!}{
    \includegraphics{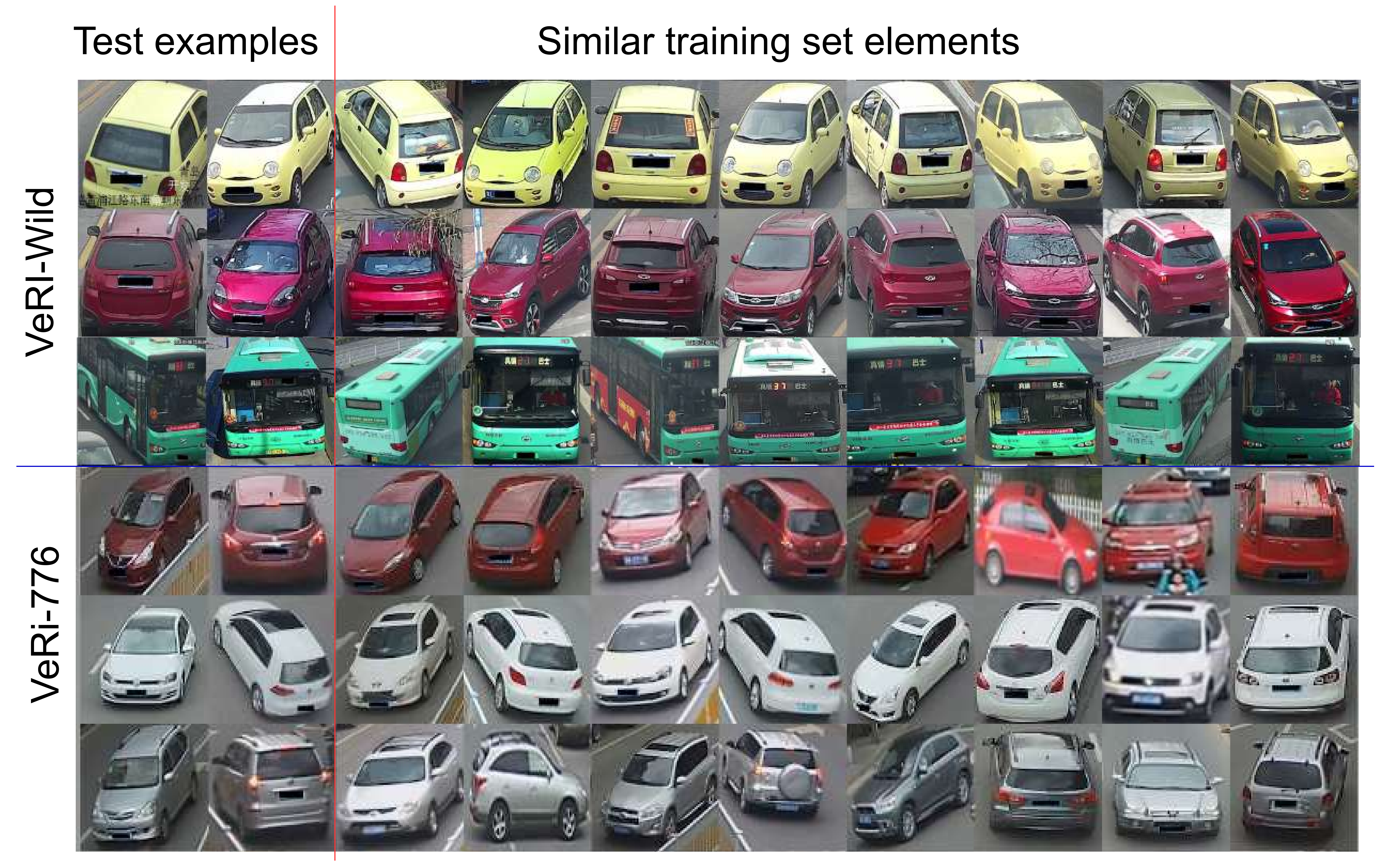}}
    \caption{Vehicle re-identification test sets contain very similar samples to those seen during training. Each row contains 5 identities, each with a frontal and rear view image pair. The first two columns show a query identity, while the remaining columns show training set samples. Examples are picked from the VeRI-WILD~\cite{Lou_2019_CVPR} and VeRi776~\cite{liu2016large} dataset.}
    \label{fig:very_similar_examples}
    \vspace{-1em}
    
\end{figure}

The most popular recent approaches to recognition and re-identification rely on deep metric learning~\cite{ye2024transformerobjectreidentificationsurvey,khan2019survey}. These approaches learn an embedding from the image space into a latent space that abides by the following rule: instances of the same object are clustered together.
When employing a standard contrastive learning approach~\cite{hermans2017defense}, the training encourages the networks to look for commonalities between positive images and discrepancies between negatives.
However, without adequate supervision, these methods tend to generalize poorly and end up finding spurious similarities and differences~\cite{kaya2019deep,khan2019survey}. 
Thus, more comprehensive datasets are needed~\cite{wang2019survey,wang2021deep}. Obtaining such datasets is often challenging due to privacy concerns, making normalization and pre-processing paramount in certain scenarios. For instance, for face recognition, landmark extraction and alignment are frequently used~\cite{lin2021deep,xu2021searching} to reduce data variance, helping to avoid modeling errors with limited available samples. 

In vehicle re-identification, we distinguish between two sub-problems. One involves comparing images of vehicles from similar viewpoints, while the other involves comparing images from different viewpoints, such as the front and back views. In fact, comparing vehicles from different view perspectives would rely on view-invariant features~\cite{li2020varid} that capture coarse details like shape and color. However, these attributes are insufficient for distinguishing between vehicles of the same color and form. 
To address this, some works have proposed incorporating view-variant and view-invariant features~\cite{bai2021disentangled}, or adding pose information~\cite{Tang_2019_ICCV} to ensure that fine-grained details are learned.
Conversely, the state-of-the-art methods~\cite{li2023ClipReidexploitingvisionlanguagemodel,he2021TransReIDtransformerbasedobjectreidentification,chen2023rotationinvarianttransformerrecognizing} for vehicle re-identification (re-ID) achieve impressive test performance on established benchmarks without relying on such pre-processing. This suggests that these methods either efficiently learn complex similarity mappings both valid across views and containing rich fine-grained information without requiring large amounts of data, or something may be amiss with the evaluation in current benchmarks.

Upon inspecting various vehicle re-ID datasets~\cite{Lou_2019_CVPR,liu2016deep}, we find that training and test samples are highly similar, as illustrated in \Cref{fig:very_similar_examples}. 
This introduces an evaluation bias towards memorizing vehicles seen during training, a task neural networks excel at~\cite{lecun2006tutorial}, regardless of whether good discriminative features are learned. To analyze this bias, we propose to split the datasets based on vehicle type, creating two identity groups -\textit{seen} and \textit{unseen} types- used to separately train and test. Our findings show that state-of-the-art methods struggle to generalize to unseen vehicle types in the standard, mixed-view, re-ID setup. Additionally, by separately considering the same and different view re-ID cases, we show that this behavior persists despite near saturated performance (97\%+) for seen types.

We detail our findings on both the VeRi776~\cite{liu2016deep} and VeRI-Wild~\cite{Lou_2019_CVPR} datasets for \textit{Seen} and \textit{Unseen} vehicle types with mixed, same, and different view evaluations.

To summarize, the contributions of this paper are: 
\begin{itemize}
    \item  We expose a bias between vehicle re-ID training and test sets and propose a new training and testing split based on vehicle type to better assess generalization.
    \item We introduce the distinction between same- and different-view re-identification as a key contribution, and split the evaluation accordingly into \textit{same-view} and \textit{different-view} settings.

\end{itemize}

\section{Related Work}
\label{sec:related_work}

\noindent {\bf Deep metric learning} seeks to learn an object representation that preserves a measure of similarity~\cite{kaya2019deep}. For that, contrastive approaches are commonly used~\cite{wang2019survey}. These approaches work by bringing samples of the same classes (positives) closer together while pushing samples of different classes (negatives) apart in a latent space. 
Initially, a cross-entropy loss~\cite{zhai2019defense} was mainly used for predicting known class labels. However, this loss does not explicitly handle the inter-class and intra-class distances. To address this, the triplet loss~\cite{hermans2017defense} was introduced, which is better for retrieval and re-identification tasks. More recently, methods such as InfoNCE~\cite{oord2018representation} loss have become popular, as they leverage multiple negatives per anchor-positive pair.
The choice of metric within these losses also varies: some use the Euclidean distance~\cite{he2018triplet}, while others use the cosine distance~\cite{deng2019arcface} to emphasize angular relationships between embeddings. Sample selection also plays an important role~\cite{xuan2020improved}, as the object viewpoint can impact the type of information learned by the network~\cite{li2020varid}.

\smallskip \noindent {\bf  Vehicle Re-Identification} is a task aimed at recognizing vehicles across multiple cameras. Early approaches built representations by combining SIFT with other representations~\cite{7553002}.
In contrast, deep learning approaches based on metric learning constitute the state of the art~\cite{khan2019survey} in vehicle re-identification. 
These approaches~\cite{zheng2019VehicleNet} use pretrained backbones, like ResNet~\cite{he2016deep} or VIT~\cite{dosovitskiy2021an}, and adapt the resulting features with a classification loss and a triplet loss.  As noted in~\cite{li2020varid}, networks do not necessarily learn relevant fine-grained information useful for distinguishing similar vehicles. For this reason, some methods propose to complement the representations with pose information~\cite{Tang_2019_ICCV} or semantic parts~\cite{shen2023triplet}, which leads to a performance improvement (up to 5\% on the VeRi776~\cite{liu2016large} dataset). 

The current state of the art is dominated by transformer-based methods~\cite{he2021TransReIDtransformerbasedobjectreidentification,li2023ClipReidexploitingvisionlanguagemodel,chen2023rotationinvarianttransformerrecognizing} such as TransReID~\cite{he2021TransReIDtransformerbasedobjectreidentification}. As a Vision Transformer~\cite{dosovitskiy2021an}, it processes images by dividing them into patches and generating a sequence of representation called tokens. An additional class token is also attached to the sequence. The sequence is then transformed by leveraging a self-attention mechanism. Similar to image classification~\cite{dosovitskiy2021an}, the class token of the output sequence is used as a representation for the current image and compared with the Euclidean distance.  TransReID further introduces a Jigsaw patch module, which  generates additional class tokens from token subgroups; these are concatenated for the final decision.
The model also incorporates viewpoint and camera information into the embedding, achieving near state-of-the-art performance. 
As mentioned in~\cite{luo2021self}, pre-training has a major impact on the final performance. Consequently, ClipReid~\cite{li2023ClipReidexploitingvisionlanguagemodel} takes advantage of the text encoder latent space of CLIP~\cite{radford2021learning} to improve the vision encoder performance. It learns text tokens to describe each vehicle identity and, in a second phase, aligns text embeddings to visual class embeddings as an additional supervision. It achieves state-of-the-art performance when using a VIT-based encoder. RotTrans~\cite{chen2023rotationinvarianttransformerrecognizing} also builds on TransReID proposing a transformer that learns rotation-robust features through a representation-level augmentation for UAV settings. 

\smallskip \noindent {\bf Vehicle Re-Identification evaluation protocols.} 
It is important to understand how the re-identification problem is currently evaluated in the literature. As proposed in~\cite{zheng2015scalable,zhu2022dual}, the evaluation task involves sorting a gallery based on similarity to multiple query images. The sorting quality is then measured using the mean average precision (mAP) and Cumulative match curve (CMC). 
The mean average precision indicates how well a method assigns low sorting ranks to the relevant matches on average. 
The CMC-$k$, on the other hand, indicates how often at least one of the matches appears before a given rank $k$.
In recent benchmarks, CMC-R1 values are often saturated, so we focus our evaluation on mAP, which provides a more discriminative measure of ranking quality.

The mean average precision ($mAP$) is computed as 
\begin{equation}
    mAP= \frac1n \sum_{q=1}^{n} AP_q, 
\end{equation}
where $n$ is the number of queried images and $AP_q$ is the average precision for a given query $q$. This average precision is calculated as
\begin{equation}
    AP_q = \frac{1}{\#Matches} \sum_{i=1}^{g} P_i \times rel_i;\quad
    P_i = \frac{TP_i}{(TP_i+FP_i)},
\end{equation}
where $P_i$ is the precision at a given rank $i$, $rel_i$ is the relevance of the $P_i$ term and it equals 1 when a positive match is at the rank $i$ and 0 otherwise, $TP_i$ ($FP_i$) is the number of positive (negative) matches present up to the rank $i$, and $\#Matches$  is the total number of matches for the query $q$ present in the gallery.
Both metrics are either reported as ratios or percentages.  
A common practice is to use re-ranking~\cite{leng2015person} to refine sorting results: it consists of re-sorting the gallery using the similarity between gallery elements. We stress that all results reported in this paper do not use re-ranking.

\begin{table} 
\caption{Vehicle Re-Identification datasets and sizes. }
\centering 
\resizebox{\linewidth}{!}{ 
\setlength{\tabcolsep}{2pt}
\begin{tabular}{c | c | c | c | c } 
dataset & id count & image count & test gallery size & number of queries\\  
\hline
VeRI-WILD~\cite{Lou_2019_CVPR} & 40,671 & 416,314 & 3k/5k/10k & 3k/5k/10k \\
VeRi 776~\cite{liu2016large} & 776 & 49,357 & 11,579 & 1678 \\ 

\hline
\end{tabular}}

\label{tab:dataset_info_vehicles}
\end{table}

We focus our studies on the VeRi776 and VeRI-Wild dataset (summarized in \Cref{tab:dataset_info_vehicles}) as they are
 the most commonly used datasets available at our disposal with VeRI-Wild being the most challenging dataset available~\cite{wang2019survey}. Both contain images of different views captured using different cameras and provide annotations for color, type, and model. Additionally, the VeRi776~\cite{liu2016large} dataset includes viewpoint annotations, while for the VeRI-Wild~\cite{Lou_2019_CVPR} dataset, we generate view annotations by training a classifier on the CityFlow~\cite{tang2019cityflow} synthetic dataset.

\section{Analysis of the Vehicle Re-Identification evaluation protocol}

\label{sec:analysis_vehicle_re_id_eval}
We start by studying the performance of  
TransReID~\cite{he2021TransReIDtransformerbasedobjectreidentification}, ClipReid~\cite{li2023ClipReidexploitingvisionlanguagemodel} and RotTrans\cite{chen2023rotationinvarianttransformerrecognizing} on both VeRI776 and VeRI-Wild. 
The evaluation in~\Cref{tab:VeRI-Wild small and VeRi776 test performance} shows that all state-of-the-art methods achieve strong performance, with mAP values exceeding 80\%. On the VeRI-Wild dataset, the performance gap between the methods is small (approximately 1\%), with ClipReid achieving the highest mAP. On the VeRI-776 dataset, this gap increases to up to 3.7\% in favor of ClipReid, which can be attributed to its CLIP-based initialization, leveraging large-scale pretraining to mitigate limited training data.
When cross-evaluating the models trained on the VeRI-Wild dataset on the VeRI776 dataset, and despite the visual similarity of vehicles across both datasets, we observe a performance drop (15\% in terms of mAP), suggesting a potential dataset bias.  Hence, we decide to closely examine the training and testing splits.

\begin{table}
\caption{Vehicle Re-Identification performance (mAP \%) on the VeRI-Wild dataset (3k) and VeRi776 dataset. Cross-dataset performance drop indicates the presence of a data bias. *Cross-dataset evaluation does not use view and camera embeddings.}
\centering 
\resizebox{\linewidth}{!}{ 
\setlength{\tabcolsep}{2pt}
\begin{tabular}{c|c|c|c}
 & VeRI-Wild & VeRi776 & {\footnotesize\shortstack{*VeRI-Wild$_{train}$\!$\to$\\[-0.5ex] VeRi776$_{test}$}} \\

\hline
TransReID\cite{he2021TransReIDtransformerbasedobjectreidentification}
& \underline{79,5} & \underline{81,3} & \underline{63,5} \\ [0.25ex]
\hline
ClipReid\cite{li2023ClipReidexploitingvisionlanguagemodel}
& \textbf{80,3} & \textbf{84,1} & 63,1 \\ [0.25ex]
\hline
RotTrans\cite{chen2023rotationinvarianttransformerrecognizing}
& 78,5 & 80,4 & \textbf{63,9} \\ [0.25ex]
\end{tabular}} 
\label{tab:VeRI-Wild small and VeRi776 test performance}
\end{table}

\subsection{Vehicle Re-identification test set bias and proposed evaluation: split with Seen Types and Unseen Types}

\label{subsec:section_2_B}

\begin{figure}
    \centering
    \includegraphics[width=\linewidth]{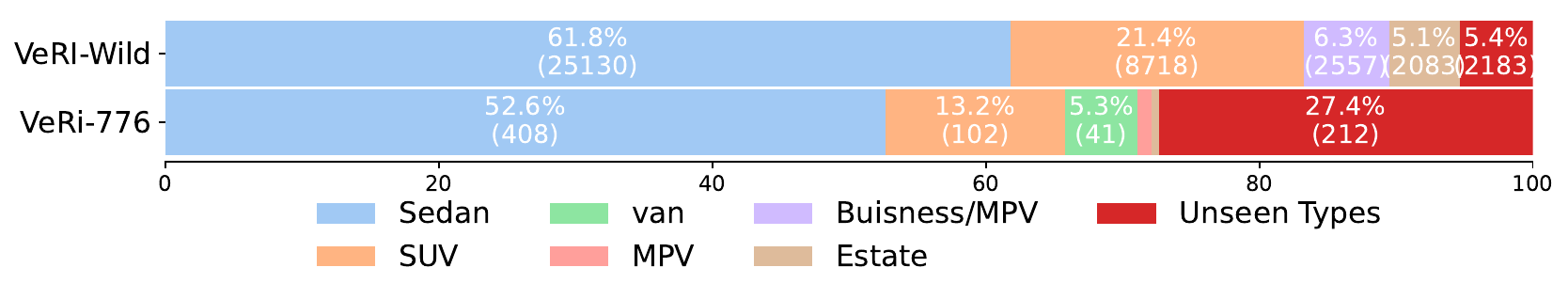}
    \caption{Proportions of unseen types per dataset. Unseen types (in red) consist of \textit{hatchbacks, buses, trucks, pick-ups and vehicles labeled as ``Others"}.}
    \label{fig:distrib_types_datasets}
\end{figure}

 Upon closer inspection of the VeRI-Wild and VeRi776 datasets, we found that the train and test splits contain many highly similar, nearly indistinguishable identities  (see \Cref{fig:very_similar_examples}), especially for overly present vehicle types such as sedans. In fact, by using the available annotation for VeRI776 and VeRI-Wild, we verify that for at least 90\% of the identities present in each test set, multiple other identities of the same type, color, and model are present in the training set. This reflects a natural bias linked to data collection, which was carried out over short periods using fixed cameras. 
 Moreover, train and test sets consist mainly of sedans as can be seen in \Cref{fig:distrib_types_datasets} (due to their commonness). This bias in the test set does not reflect the full range of real-world conditions. 

Hence, we propose slicing\footnote{\url{https://github.com/deejey674/Generalisation-limits-in-Vehicle-Re-ID/}} the dataset based on vehicle type to better assess performance. We create two identity groups based on vehicle types: \textit{Seen Types} and \textit{Unseen Types}. Seen Types are vehicle types that are used for both training and testing, while Unseen Types are reserved for testing. 
We sample queries and gallery images for both groups, selecting frontal query images and  4 matches per identity for gallery images such that half are frontal images and half are rear images.
This Mixed View setup aligns closely with the conditions of the standard evaluation.
For both the VeRi776 and VeRi-Wild datasets, the Seen Types are: \textit{sedans, MPVs, vans, SUVs, and station wagons}; with the remaining vehicle types (\textit{hatchbacks, buses, trucks, pick-ups and vehicles labeled as ``Others"}) classified as Unseen Types.

\begin{table}[t]
\caption{mAP\% of methods trained on the Seen Types set and evaluated on the \textbf{Mixed View (Standard)} galleries of Seen Types and Unseen types test sets for the VeRI-Wild~\cite{Lou_2019_CVPR} and VeRi-776~\cite{liu2016large} datasets. The best results along a given column are highlighted in bold.}
    \label{tab:VeRi776_MAP_CMC_Mixed_view}
    \centering
    \scalebox{1.2}{
    \setlength{\tabcolsep}{2pt}
    \begin{tabular}{c|c|c||c}
          &&Seen types&Unseen types\\
          \cline{3-4}
          TransReID~\cite{he2021TransReIDtransformerbasedobjectreidentification}&\multirow{3}{*}{\rotatebox[origin=c]{90}{\fontsize{6}{9}\selectfont\hspace{1.5em}VeRI-Wild}}&  98,6 &  65,7  \\
          ClipReid~\cite{li2023ClipReidexploitingvisionlanguagemodel} &&  \underline{98.7} & \underline{67,7} \\
          RotTrans~\cite{chen2023rotationinvarianttransformerrecognizing} &&  \textbf{99,1}& \textbf{70}\\
 
          \hline\hline
          
          TransReID~\cite{he2021TransReIDtransformerbasedobjectreidentification} 
&\multirow{3}{*}{\rotatebox[origin=c]{90}{\fontsize{6}{9}\selectfont\hspace{-0.5em}VeRi776}}& 90,4& 57,2 \\
 ClipReid~\cite{li2023ClipReidexploitingvisionlanguagemodel} 
& & \underline{91} &\textbf{60,2 }\\
 RotTrans~\cite{chen2023rotationinvarianttransformerrecognizing} 
& & \textbf{91,3} & \underline{58,5} \\

\end{tabular}}

\end{table}

We retrain TransReID, ClipReid, and RotTrans following the guidelines of their respective papers for both datasets.

The results in \Cref{tab:VeRi776_MAP_CMC_Mixed_view} show that, for vehicle types seen during training (Seen Types), ClipReid, TransReID, and RotTrans all exceed the 98\% mark for VeRI-Wild and 90\% mark for VeRi776, in terms of mAP, with a relatively small performance gap between them (up to 0.4\%). This shows that the state-of-the-art models can reliably match elements of different views for known type. 

However, when moving to the Unseen Types, we see that the performance gap grows between the different methods (from 0,4\% to 2.3\% between ClipReid and RotTrans on the VeRI-Wild dataset). We observe that, for the larger dataset VeRI-Wild, RotTrans is more robust to view changes than ClipReid, but for the smaller dataset VeRI776, the latter performs better by leveraging the latent space of Clip. Most importantly, we observe a substantial performance drop for all methods of up to 30\% in terms of mAP. This shows that networks struggle to generalize to unseen vehicle types. 

As the re-ID task requires both attention to details and viewpoint robustness and due to the mixed nature of the gallery in terms of viewpoints, it is difficult to infer which sub-task generalizes poorly.
For this purpose, we introduce the distinction between same- and diffrent -view re-identification.

\begin{table}[t]
\caption{mAP\% of methods trained on the Seen Types set and evaluated on the \textbf{Same View} galleries of Seen Types and Unseen types test sets for the VeRI-Wild~\cite{Lou_2019_CVPR} and VeRi-776~\cite{liu2016large} datasets. The best results along a given column are highlighted in bold.}
    \label{tab:VeRi776_MAP_CMC_Same_View}
    \centering
    \scalebox{1.25}{
    \setlength{\tabcolsep}{2pt}
    \begin{tabular}{c|c|c|c|}
          &&  Seen types&  Unseen types\\
          \cline{3-4}
          TransReID~\cite{he2021TransReIDtransformerbasedobjectreidentification}&\multirow{3}{*}{\rotatebox[origin=c]{90}{\fontsize{6}{9}\selectfont\hspace{1.5em}VeRI-Wild}}&  99,1 & \underline{80,4} \\
          ClipReid~\cite{li2023ClipReidexploitingvisionlanguagemodel} &&  \textbf{99,8}   &  79,6 \\
          RotTrans~\cite{chen2023rotationinvarianttransformerrecognizing} &&  \underline{99,2} &  \textbf{84,2} \\
 
          \hline\hline
          
          TransReID~\cite{he2021TransReIDtransformerbasedobjectreidentification} 
&\multirow{3}{*}{\rotatebox[origin=c]{90}{\fontsize{6}{9}\selectfont\hspace{-0.5em}VeRi776}}&  93,1  &  66,5 \\
 ClipReid~\cite{li2023ClipReidexploitingvisionlanguagemodel} 
& & \underline{93,8}  & \textbf{69,4}\\
 RotTrans~\cite{chen2023rotationinvarianttransformerrecognizing} 
& & \textbf{94,9} & \underline{68,8} \\

    \end{tabular}}

\end{table}

\begin{table}[t]
\caption{mAP\% of methods trained on the Seen Types set and evaluated on the \textbf{Different View } galleries of Seen Types and Unseen types test sets for the VeRI-Wild~\cite{Lou_2019_CVPR} and VeRi-776~\cite{liu2016large} datasets. The best results along a given column are highlighted in bold.}
    \label{tab:VeRi776_MAP_CMC_diff_view}
    \centering
    \scalebox{1.2}{
    \setlength{\tabcolsep}{2pt}
    \begin{tabular}{c|c|c||c}
          &&Seen types&Unseen types\\
          \cline{3-4}
          TransReID~\cite{he2021TransReIDtransformerbasedobjectreidentification}&\multirow{3}{*}{\rotatebox[origin=c]{90}{\fontsize{6}{9}\selectfont\hspace{1.5em}VeRI-Wild}}&  96,9&    50,2\\
          ClipReid~\cite{li2023ClipReidexploitingvisionlanguagemodel} &&  \underline{97,1}&  \underline{51,8}\\
          RotTrans~\cite{chen2023rotationinvarianttransformerrecognizing} &&  \textbf{97,9}& \textbf{56,6}\\
 
          \hline\hline
          
          TransReID~\cite{he2021TransReIDtransformerbasedobjectreidentification} 
&\multirow{3}{*}{\rotatebox[origin=c]{90}{\fontsize{6}{9}\selectfont\hspace{-0.5em}VeRi776}}& \underline{89,7}&  \textbf{46,6}\\
 ClipReid~\cite{li2023ClipReidexploitingvisionlanguagemodel} 
& &  \textbf{91.6}& \underline{44.9}\\
 RotTrans~\cite{chen2023rotationinvarianttransformerrecognizing} 
& & 88,9&  44,5\\

\end{tabular}}

\end{table}

\subsection{Same and different view Re-Identification}
\label{section_2_A}

The standard re-identification evaluation protocol~\cite{zheng2015scalable,zhu2022dual} involves sorting a gallery of images in terms of similarity to the query image in search of matches.  Both query and gallery images contain a mix of frontal and rear views. Depending on the gallery-query image pair, we can distinguish two cases:
\begin{itemize}
    \item \textit{\textbf{Same-view Re-Identification}}. The two viewing angles are similar, e.g. front-front or rear-rear comparisons. Accomplishing this sub-task requires attention to details such as stickers, scratches, or logos.  
    \item \textit{\textbf{Different-view Re-Identification}}. The two viewing angles are so dissimilar that there is no overlap between the views, e.g. front-rear comparisons. This requires using prior knowledge to extrapolate information or relying on view-invariant information such as color.  
\end{itemize}

\begin{figure}[t]
     \centering
     \resizebox{\linewidth}{!}{ 
     \includegraphics{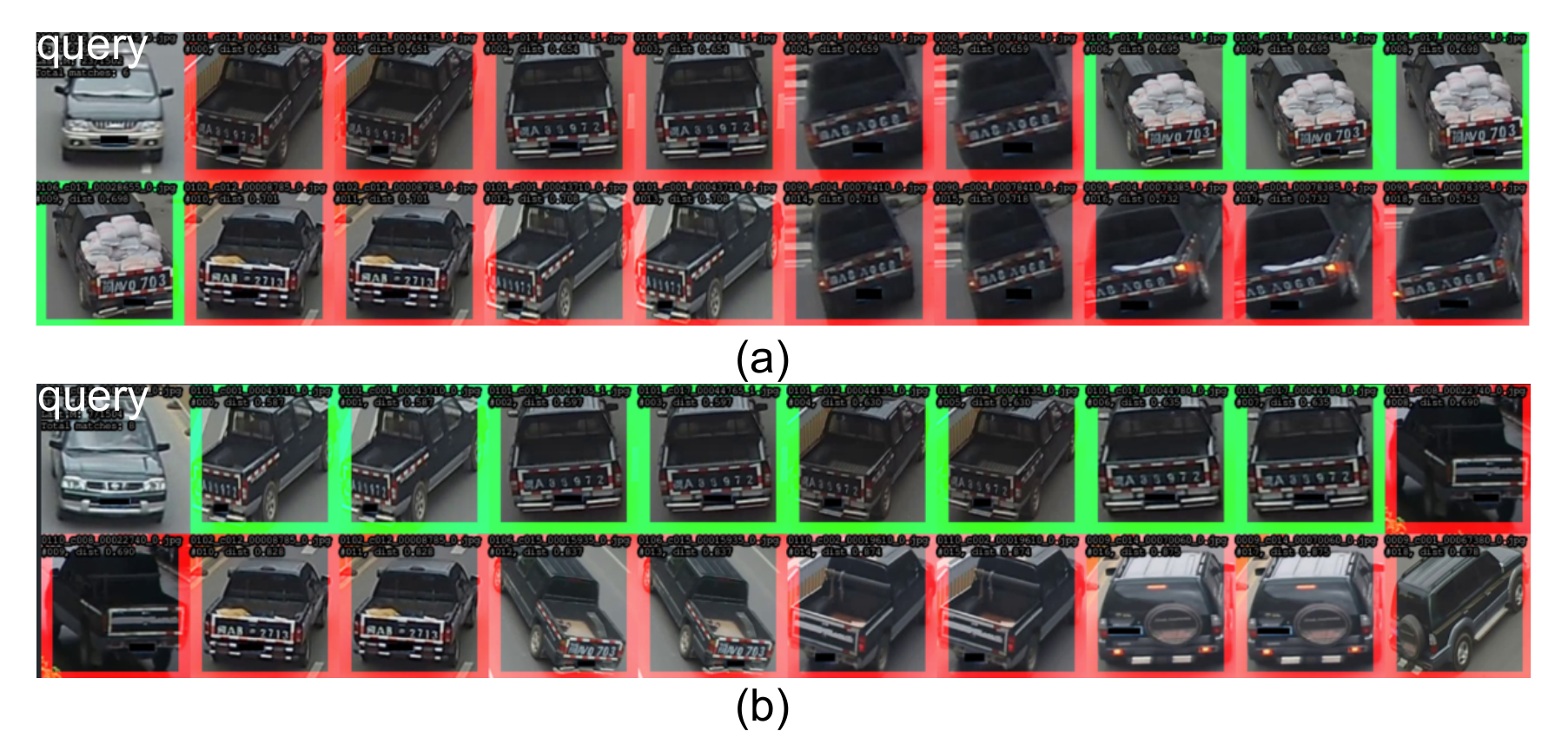}
     }
     \caption{Different sorted gallery examples obtained with TransReID~\cite{he2021TransReIDtransformerbasedobjectreidentification} on VeRi776~\cite{liu2016large}. All cases show a front view query image compared to rear view images. Case (a) shows bad matches (highlighted in red) ranked before good ones (green) while case (b) shows the opposite. Differentiating between correct and wrong matches requires rear-view information that is absent from the query image. The presence or absence of the cargo cannot be verified from the front-view query image. }
     \label{fig:Front_q_rear_Gallery}
 \end{figure}

Vehicle re-identification across different views is inherently ambiguous due to a lack of common visual information~\Cref{fig:Front_q_rear_Gallery}. This presents a distinct problem that might require different techniques compared to same-view re-identification. Therefore, mixing the evaluation of these two tasks can be misleading. 
For this purpose, for both the seen and unseen type identity sets, we create two new galleries: Same View and Different view galleries. This is achieved by swapping out rear-view images to front-view ones from the mixed view gallery for same-view and vice-versa for different view.

As seen in \Cref{tab:VeRi776_MAP_CMC_Same_View}, for vehicle types seen during training (Seen Types), ClipReid, TransReID, and RotTrans all exceed the 99\% mark in terms of mAP for the same-view evaluation on VeRI-Wild and the 93\% mark for VeRi776.  
When moving to the Unseen Types, we see higher performance for the same-view setup relative to the standard mixed view re-ID setup, but still observe a substantial performance drop for all methods of nearly 20\% compared to the seen types performance. This shows that, for the same-view setting, networks fail to handle unseen vehicle types well.  
Additionally, when looking at the different-view evaluation in \Cref{tab:VeRi776_MAP_CMC_diff_view}, we see high performance (near 90\%) for most methods for seen types, with a gap of up to 5\% relative to same view performance, indicating that the task is harder. This gap increases to 30\% at least, relative to the same-view evaluation on unseen types, highlighting a difficulty to maintain viewpoint robustness for unseen vehicle types.
This exposes a flaw as networks struggle to generalize to unseen vehicle types in both same and different view re-ID.

\section{Conclusion}
\label{sec:conclusion}

In this paper, we showcase a bias in vehicle re-identification train and test sets of VeRI776 and VeRI-Wild. We address this issue by proposing an alternative evaluation protocol that more effectively measures generalization capability to unseen vehicle types. Furthermore, we distinguish between two sub-tasks: same-view and different-view re-identification, demonstrating that state-of-the-art methods struggle to generalize to unseen vehicle types for both re-ID settings.
\\

{\bf Acknowledgments}: This work was granted access to
the HPC resources of IDRIS under the allocation 20XXAD011013861R2
made by GENCI



{
    \small
    \bibliographystyle{IEEEtran}
    \bibliography{IEEEabrv}

\begin{thebibliography}{10}
\providecommand{\url}[1]{#1}
\csname url@samestyle\endcsname
\providecommand{\newblock}{\relax}
\providecommand{\bibinfo}[2]{#2}
\providecommand{\BIBentrySTDinterwordspacing}{\spaceskip=0pt\relax}
\providecommand{\BIBentryALTinterwordstretchfactor}{4}
\providecommand{\BIBentryALTinterwordspacing}{\spaceskip=\fontdimen2\font plus
\BIBentryALTinterwordstretchfactor\fontdimen3\font minus
  \fontdimen4\font\relax}
\providecommand{\BIBforeignlanguage}[2]{{%
\expandafter\ifx\csname l@#1\endcsname\relax
\typeout{** WARNING: IEEEtran.bst: No hyphenation pattern has been}%
\typeout{** loaded for the language `#1'. Using the pattern for}%
\typeout{** the default language instead.}%
\else
\language=\csname l@#1\endcsname
\fi
#2}}
\providecommand{\BIBdecl}{\relax}
\BIBdecl

\bibitem{5204071}
T.~Hou, S.~Wang, and H.~Qin, ``Vehicle matching and recognition under large
  variations of pose and illumination,'' in \emph{2009 IEEE COMP SOC ANN},
  2009, pp. 24--29.

\bibitem{Sun_2019_CVPR}
X.~Sun and L.~Zheng, ``Dissecting person re-identification from the viewpoint
  of viewpoint,'' in \emph{Proc. IEEE/CVF Conf. Comput. Vis. Pattern Recognit.
  (CVPR)}, June 2019.

\bibitem{li2020infrared}
D.~Li, X.~Wei, X.~Hong, and Y.~Gong, ``Infrared-visible cross-modal person
  re-identification with an x modality,'' in \emph{Proc. AAAI Conf. Artif.
  Intell.}, vol.~34, no.~04, 2020, pp. 4610--4617.

\bibitem{steger2002occlusion}
C.~Steger, ``Occlusion, clutter, and illumination invariant object
  recognition,'' \emph{Int. Arch. Photogramm. Remote Sens. Spatial Inf. Sci.},
  vol.~34, no. 3/A, pp. 345--350, 2002.

\bibitem{8283763}
W.~Zhang, X.~Zhao, J.-M. Morvan, and L.~Chen, ``Improving shadow suppression
  for illumination robust face recognition,'' \emph{IEEE TR PARTS MATER},
  vol.~41, no.~3, pp. 611--624, 2019.

\bibitem{koo2022survey}
J.~H. Koo, S.~W. Cho, N.~R. Baek, Y.~W. Lee, and K.~R. Park, ``A survey on face
  and body based human recognition robust to image blurring and low
  illumination,'' \emph{Mathematics}, vol.~10, no.~9, p. 1522, 2022.

\bibitem{wang2019survey}
H.~Wang, J.~Hou, and N.~Chen, ``A survey of vehicle re-identification based on
  deep learning,'' \emph{IEEE Access}, vol.~7, pp. 172\,443--172\,469, 2019.

\bibitem{wang2021deep}
M.~Wang and W.~Deng, ``Deep face recognition: A survey,''
  \emph{Neurocomputing}, vol. 429, pp. 215--244, 2021.

\bibitem{ye2021deep}
M.~Ye, J.~Shen, G.~Lin, T.~Xiang, L.~Shao, and S.~C. Hoi, ``Deep learning for
  person re-identification: A survey and outlook,'' \emph{IEEE TR PARTS MATER},
  vol.~44, no.~6, pp. 2872--2893, 2021.

\bibitem{ye2024transformerobjectreidentificationsurvey}
M.~Ye, S.~Chen, C.~Li, W.-S. Zheng, D.~Crandall, and B.~Du, ``Transformer for
  object re-identification: A survey,'' \emph{International Journal of Computer
  Vision}, pp. 1--31, 2024.

\bibitem{Cermak_2024_WACV}
V.~\v{C}erm\'ak, L.~Picek, L.~Adam, and K.~Papafitsoros, ``Wildlifedatasets: An
  open-source toolkit for animal re-identification,'' in \emph{Proc. IEEE/CVF
  Winter Conf. Appl. Comput. Vis. (WACV)}, January 2024, pp. 5953--5963.

\bibitem{khan2019survey}
S.~D. Khan and H.~Ullah, ``A survey of advances in vision-based vehicle
  re-identification,'' \emph{Comput. Vis. Image Underst.}, vol. 182, pp.
  50--63, 2019.

\bibitem{Lou_2019_CVPR}
Y.~Lou, Y.~Bai, J.~Liu, S.~Wang, and L.~Duan, ``Veri-wild: A large dataset and
  a new method for vehicle re-identification in the wild,'' in \emph{Proc.
  IEEE/CVF Conf. Comput. Vis. Pattern Recognit. (CVPR)}, June 2019.

\bibitem{liu2016large}
X.~Liu, W.~Liu, H.~Ma, and H.~Fu, ``Large-scale vehicle re-identification in
  urban surveillance videos,'' in \emph{2016 IEEE Int. Conf. Multimedia Expo
  (ICME)}.\hskip 1em plus 0.5em minus 0.4em\relax IEEE, 2016, pp. 1--6.

\bibitem{hermans2017defense}
A.~Hermans, L.~Beyer, and B.~Leibe, ``In defense of the triplet loss for person
  re-identification,'' \emph{arXiv preprint arXiv:1703.07737}, 2017.

\bibitem{kaya2019deep}
M.~Kaya and H.~{\c{S}}. Bilge, ``Deep metric learning: A survey,''
  \emph{Symmetry}, vol.~11, no.~9, p. 1066, 2019.

\bibitem{lin2021deep}
C.-H. Lin, W.-J. Huang, and B.-F. Wu, ``Deep representation alignment network
  for pose-invariant face recognition,'' \emph{Neurocomputing}, vol. 464, pp.
  485--496, 2021.

\bibitem{xu2021searching}
X.~Xu, Q.~Meng, Y.~Qin, J.~Guo, C.~Zhao, F.~Zhou, and Z.~Lei, ``Searching for
  alignment in face recognition,'' in \emph{Proc. AAAI Conf. Artif. Intell.},
  vol.~35, no.~4, 2021, pp. 3065--3073.

\bibitem{li2020varid}
Y.~Li, K.~Liu, Y.~Jin, T.~Wang, and W.~Lin, ``Varid: Viewpoint-aware
  re-identification of vehicle based on triplet loss,'' \emph{IIEEE Trans.
  Intell. Transp. Syst.}, vol.~23, no.~2, pp. 1381--1390, 2020.

\bibitem{bai2021disentangled}
Y.~Bai, J.~Liu, Y.~Lou, C.~Wang, and L.-Y. Duan, ``Disentangled feature
  learning network and a comprehensive benchmark for vehicle
  re-identification,'' \emph{IEEE TR PARTS MATER}, vol.~44, no.~10, pp.
  6854--6871, 2021.

\bibitem{Tang_2019_ICCV}
Z.~Tang, M.~Naphade, S.~Birchfield, J.~Tremblay, W.~Hodge, R.~Kumar, S.~Wang,
  and X.~Yang, ``Pamtri: Pose-aware multi-task learning for vehicle
  re-identification using highly randomized synthetic data,'' in \emph{Proc.
  IEEE/CVF Int. Conf. Comput. Vis. (ICCV)}, October 2019.

\bibitem{li2023ClipReidexploitingvisionlanguagemodel}
S.~Li, L.~Sun, and Q.~Li, ``Clip-reid: Exploiting vision-language model for
  image re-identification without concrete text labels,'' \emph{arXiv preprint
  arXiv:2211.13977}, 2022.

\bibitem{he2021TransReIDtransformerbasedobjectreidentification}
S.~He, H.~Luo, P.~Wang, F.~Wang, H.~Li, and W.~Jiang, ``Transreid:
  Transformer-based object re-identification,'' in \emph{Proc. IEEE/CVF Int.
  Conf. Comput. Vis. (ICCV)}, October 2021, pp. 15\,013--15\,022.

\bibitem{chen2023rotationinvarianttransformerrecognizing}
S.~Chen, M.~Ye, and B.~Du, ``Rotation invariant transformer for recognizing
  object in uavs,'' in \emph{Proc. 30th ACM Int. Conf. Multimedia}, 2022, pp.
  2565--2574.

\bibitem{liu2016deep}
X.~Liu, W.~Liu, T.~Mei, and H.~Ma, ``A deep learning-based approach to
  progressive vehicle re-identification for urban surveillance,'' in
  \emph{Proc. European Conf. Comput. Vis. (ECCV)}.\hskip 1em plus 0.5em minus
  0.4em\relax Springer, 2016, pp. 869--884.

\bibitem{lecun2006tutorial}
Y.~LeCun, S.~Chopra, R.~Hadsell, M.~Ranzato, and F.~Huang, ``A tutorial on
  energy-based learning,'' \emph{Predicting structured data}, vol.~1, no.~0,
  2006.

\bibitem{zhai2019defense}
Y.~Zhai, X.~Guo, Y.~Lu, and H.~Li, ``In defense of the classification loss for
  person re-identification,'' in \emph{Proc. IEEE/CVF Conf. Comput. Vis.
  Pattern Recognit. (CVPR) Workshops}, 2019, pp. 0--0.

\bibitem{oord2018representation}
A.~v.~d. Oord, Y.~Li, and O.~Vinyals, ``Representation learning with
  contrastive predictive coding,'' \emph{arXiv preprint arXiv:1807.03748},
  2018.

\bibitem{he2018triplet}
X.~He, Y.~Zhou, Z.~Zhou, S.~Bai, and X.~Bai, ``Triplet-center loss for
  multi-view 3d object retrieval,'' in \emph{Proc. IEEE Conf. Comput. Vis.
  Pattern Recognit. (CVPR)}, 2018, pp. 1945--1954.

\bibitem{deng2019arcface}
J.~Deng, J.~Guo, N.~Xue, and S.~Zafeiriou, ``Arcface: Additive angular margin
  loss for deep face recognition,'' in \emph{Proc. IEEE/CVF Conf. Comput. Vis.
  Pattern Recognit. (CVPR)}, 2019, pp. 4690--4699.

\bibitem{xuan2020improved}
H.~Xuan, A.~Stylianou, and R.~Pless, ``Improved embeddings with easy positive
  triplet mining,'' in \emph{Proc. IEEE/CVF Winter Conf. Appl. Comput. Vis.
  (WACV)}, 2020, pp. 2474--2482.

\bibitem{7553002}
X.~Liu, W.~Liu, H.~Ma, and H.~Fu, ``Large-scale vehicle re-identification in
  urban surveillance videos,'' in \emph{2016 IEEE Int. Conf. Multimedia Expo
  (ICME)}, 2016, pp. 1--6.

\bibitem{zheng2019VehicleNet}
Z.~Zheng, T.~Ruan, Y.~Wei, and Y.~Yang, ``Vehiclenet: Learning robust feature
  representation for vehicle re-identification.'' in \emph{Proc. IEEE Conf.
  Comput. Vis. Pattern Recognit. (CVPR) Workshops}, vol.~2, no.~3, 2019.

\bibitem{he2016deep}
K.~He, X.~Zhang, S.~Ren, and J.~Sun, ``Deep residual learning for image
  recognition,'' in \emph{Proc. IEEE Conf. Comput. Vis. Pattern Recognit.
  (CVPR)}, 2016, pp. 770--778.

\bibitem{dosovitskiy2021an}
\BIBentryALTinterwordspacing
A.~Dosovitskiy, L.~Beyer, A.~Kolesnikov, D.~Weissenborn, X.~Zhai,
  T.~Unterthiner, M.~Dehghani, M.~Minderer, G.~Heigold, S.~Gelly, J.~Uszkoreit,
  and N.~Houlsby, ``An image is worth 16x16 words: Transformers for image
  recognition at scale,'' in \emph{Int. Conf. on Learn. Represent. (ICLR)},
  2021. [Online]. Available: \url{https://openreview.net/forum?id=YicbFdNTTy}
\BIBentrySTDinterwordspacing

\bibitem{shen2023triplet}
F.~Shen, X.~Du, L.~Zhang, X.~Shu, and J.~Tang, ``Triplet contrastive
  representation learning for unsupervised vehicle re-identification,''
  \emph{arXiv preprint arXiv:2301.09498}, 2023.

\bibitem{luo2021self}
H.~Luo, P.~Wang, Y.~Xu, F.~Ding, Y.~Zhou, F.~Wang, H.~Li, and R.~Jin,
  ``Self-supervised pre-training for transformer-based person
  re-identification,'' \emph{arXiv preprint arXiv:2111.12084}, 2021.

\bibitem{radford2021learning}
A.~Radford, J.~W. Kim, C.~Hallacy, A.~Ramesh, G.~Goh, S.~Agarwal, G.~Sastry,
  A.~Askell, P.~Mishkin, J.~Clark \emph{et~al.}, ``Learning transferable visual
  models from natural language supervision,'' in \emph{Int. Conf. on machine
  learning (ICML)}.\hskip 1em plus 0.5em minus 0.4em\relax PMLR, 2021, pp.
  8748--8763.

\bibitem{zheng2015scalable}
L.~Zheng, L.~Shen, L.~Tian, S.~Wang, J.~Wang, and Q.~Tian, ``Scalable person
  re-identification: A benchmark,'' in \emph{Proc. IEEE Int. Conf. Comput. Vis.
  (ICCV)}, 2015, pp. 1116--1124.

\bibitem{zhu2022dual}
H.~Zhu, W.~Ke, D.~Li, J.~Liu, L.~Tian, and Y.~Shan, ``Dual cross-attention
  learning for fine-grained visual categorization and object
  re-identification,'' in \emph{Proc. IEEE/CVF Conf. Comput. Vis. Pattern
  Recognit. (CVPR)}, 2022, pp. 4692--4702.

\bibitem{leng2015person}
Q.~Leng, R.~Hu, C.~Liang, Y.~Wang, and J.~Chen, ``Person re-identification with
  content and context re-ranking,'' \emph{Multimedia Tools and Applications},
  vol.~74, pp. 6989--7014, 2015.

\bibitem{tang2019cityflow}
Z.~Tang, M.~Naphade, M.-Y. Liu, X.~Yang, S.~Birchfield, S.~Wang, R.~Kumar,
  D.~Anastasiu, and J.-N. Hwang, ``Cityflow: A city-scale benchmark for
  multi-target multi-camera vehicle tracking and re-identification,'' in
  \emph{Proc. IEEE/CVF Conf. Comput. Vis. Pattern Recognit. (CVPR)}, 2019, pp.
  8797--8806.

\end{thebibliography}
}

\vspace{12pt}
\color{red}

\end{document}